\documentclass[letterpaper, 10 pt, conference]{ieeeconf}  

\usepackage{textcomp}
\usepackage{amsfonts,amssymb} 
\usepackage{booktabs}
\usepackage{stfloats}
\usepackage{url}
\usepackage{verbatim}
\usepackage{graphicx}
\usepackage{cite}
\usepackage[version=4]{mhchem}
\usepackage{graphicx} 
\usepackage{epsfig} 
\usepackage{mathptmx} 
\usepackage{times} 
\usepackage{amsmath} 
\usepackage{amssymb}  
\usepackage[T1]{fontenc}
\usepackage{tabularx}  
\usepackage{hyperref}
\hypersetup{hidelinks,
	colorlinks=true,
	allcolors=black,
	pdfstartview=Fit,
	breaklinks=true}
\usepackage{multirow}
\usepackage{url}
\usepackage{makecell}
\usepackage{diagbox} 
\usepackage{bm}
\usepackage[linesnumbered,ruled,vlined]{algorithm2e}

\IEEEoverridecommandlockouts                              
\title{\LARGE \bf
Visuo-Tactile Sensor Enabled Pneumatic Device Towards Compliant Oropharyngeal Swab Sampling
}

\author{Shoujie Li, Mingshan He, Wenbo Ding, Linqi Ye, Xueqian Wang, Junbo Tan, Jinqiu Yuan,\\ Xiao-Ping Zhang,~\IEEEmembership{Fellow,~IEEE}
\thanks{This work is supported in part by the grant from the Institute for Guo Qiang of Tsinghua University 2020GQG1004, by Tsinghua Shenzhen International Graduate School (QD2021013C, QD2022009C). (Corresponding author: Wenbo Ding, ding.wenbo@sz.tsinghua.edu.cn) }
\thanks{Shoujie Li, Wenbo Ding, Xueqian Wang, Junbo Tan, and Xiao-Ping Zhang are with Tsinghua Shenzhen International Graduate School, 518055 Shenzhen, China.}
\thanks{Mingshan He is with the Department of Mechanical Engineering, Seoul National University, Seoul, Korea.}%
\thanks{Linqi Ye is with the Department of Intelligent Control, Qianyuan Institute of Science, 310024 Hangzhou, China.}
\thanks{Jinqiu Yuan is Clinical Research Center, The Seventh Affiliated Hospital, Sun Yat-sen University, Shenzhen, Guangdong, 518107, China.}
\thanks{Xiao-Ping Zhang is also with the Department of Electrical, Computer and Biomedical Engineering, Ryerson University, Toronto, ON M5B 2K3, Canada.}
}

\begin{document}

\maketitle
\thispagestyle{empty}
\pagestyle{empty}

\begin{abstract}
Manual oropharyngeal (OP) swab sampling is an intensive and risky task. In this article, a novel OP swab sampling device of low cost and high compliance is designed by combining the visuo-tactile sensor and the pneumatic actuator-based gripper. Here, a concave visuo-tactile sensor called CoTac is first proposed to address the problems of high cost and poor reliability of traditional multi-axis force sensors. Besides, 
by imitating the doctor's fingers, a soft pneumatic actuator with a rigid skeleton structure is designed, which is demonstrated to be reliable and safe via finite element modeling and experiments. Furthermore, we propose a sampling method that adopts a compliant control algorithm based on the adaptive virtual force to enhance the safety and compliance of the swab sampling process. The effectiveness of the device has been verified through sampling experiments as well as in vivo tests, indicating great application potential. The cost of the device is around 30 US dollars and the total weight of the functional part is  less than  0.1 kg, allowing the device to be rapidly deployed on various robotic arms.  Videos, hardware, and source code are available at: https://sites.google.com/view/swab-sampling/.
\end{abstract}

\section{Introduction}
The explosive spread and rampage of the Corona Virus Disease 2019 (COVID-19) since 2019 has caused more than 700 million infections and 6 million deaths until February 2023 and radically changed human society in many aspects~\cite{Dashboard}. Timely and accurate swab sampling tests have proved to be the most efficient method to monitor and suppress the spread of the virus before the invention and validation of effective vaccines. Currently, the swab sampling methods of initial diagnostic testing for COVID-19 mainly include the nasopharyngeal (NP) swab~\cite{wang2020design,maeng2022development} and the oropharyngeal (OP) swab~\cite{yang2020combating}. Considering the comfort and convenience, OP swab sampling is more popular and has become mainstream in daily regular tests. However, the frequent sampling for a large population still brings a huge burden to both the physical and mental health of the medical staff \cite{al2022immediate}.  In addition to COVID-19, pharyngeal swab sampling plays an important role in the diagnosis of influenza and respiratory viruses \cite{galli2020self,ali2015throat}.  To this end, designing a safe and efficient OP swab sampling robot is of great significance and urgently needed.

\begin{figure}
	\centering
	\includegraphics[width=0.48\textwidth]{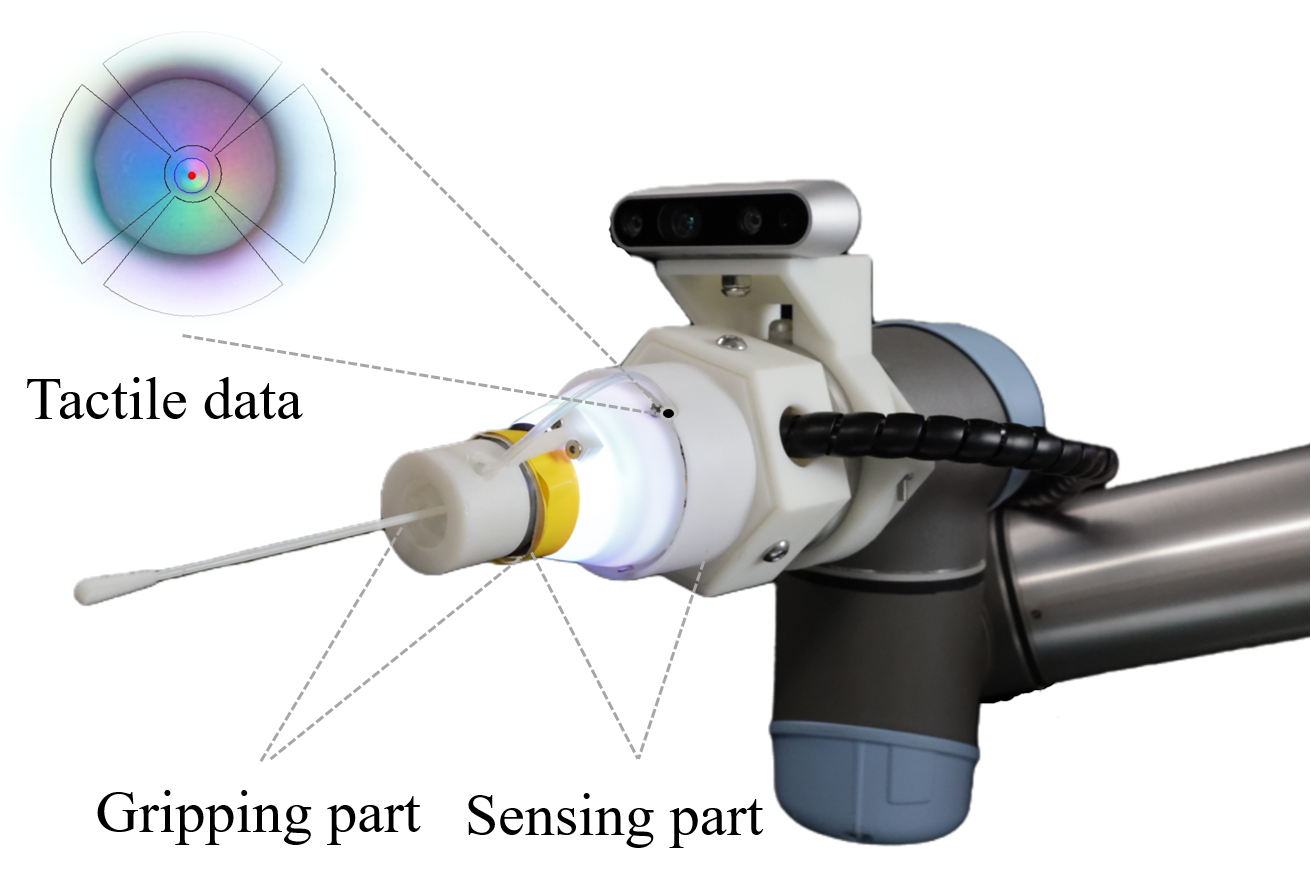}
	\caption{The proposed OP swab sampling device is based on visuo-tactile sensor and pneumatic gripper.} \label{fig:device}
\end{figure}

At the very beginning, the OP swab sampling robots were designed and remotely controlled by the medical staff to avoid direct contact with the patients. Representative works include the remote sampling system with a flexible front-end structure~\cite{li2021flexible}, the teleoperated swab robot enabled by the soft hand~\cite{chen2021tele}, the semi-automatic OP swab sampling robot~\cite{li2020clinical}, etc. Unfortunately, as the sampling amount increases, the throughput and efficiency of such semi-automatic robots have limited their applications. Consequently, the OP sampling systems based on the commercial fully-automated robots have been designed, such as the automated robot invented by Lifeline Robotics~\cite{medgadget}, and the highly efficient auto throat swab robots from Tsinghua University~\cite{globaltimes}.
It should be noted that the system design based on commercial robots is of great stability and reliability while lacking consideration for softness and comfort. To tackle this issue, Hu \textit{et al.} proposed a rigid-flexible coupled automated OP swab sampling device with both motorized and pneumatic methods, which has better compliance~\cite{hu2021design}. Nevertheless, the structure of this robot is complicated and some of its components are expensive. Therefore, it is of great meaning to design a safe OP swab sampling robot with integrated automation, compliance, simple structure, and low cost.

The safety of the OP swab sampling robots highly relies on tactile sensors and compliant control algorithms. Traditional tactile perception techniques based on different electro-mechanical conversion mechanisms~\cite{tiwana2012review} usually have a zero drift and require continuous calibration during operation. Moreover, the complex peripheral reading circuits make it difficult and expensive to realize large-area, high-resolution, and accurate tactile sensing. Thanks to the commercialization and miniaturization of the CMOS image chips, the visuo-tactile sensors~\cite{lambeta2020digit,yuan2017gelsight,9811806} have emerged as a game breaker for tactile perception, which could easily achieve high resolution with low circuit complexity. In addition to the tactile sensor design, robust and compliant control algorithms ~\cite{VANDERBORGHT20131601,7384497} are the other critical factor for the safety of the sampling robot~\cite{app10196923,8764016}. Classical compliant control methods usually have fixed coefficients, and might not be directly applicable to the OP swab sampling which is a long-term and high-intensity task under highly dynamic environments. The adaptive control strategy can adjust the parameters in real-time through self-adaptation, which not only increases compliance in the sampling process but also improves the robustness of the control~\cite{7330025}.

\begin{figure}[h]
	\centering
	\includegraphics[width=0.48\textwidth]{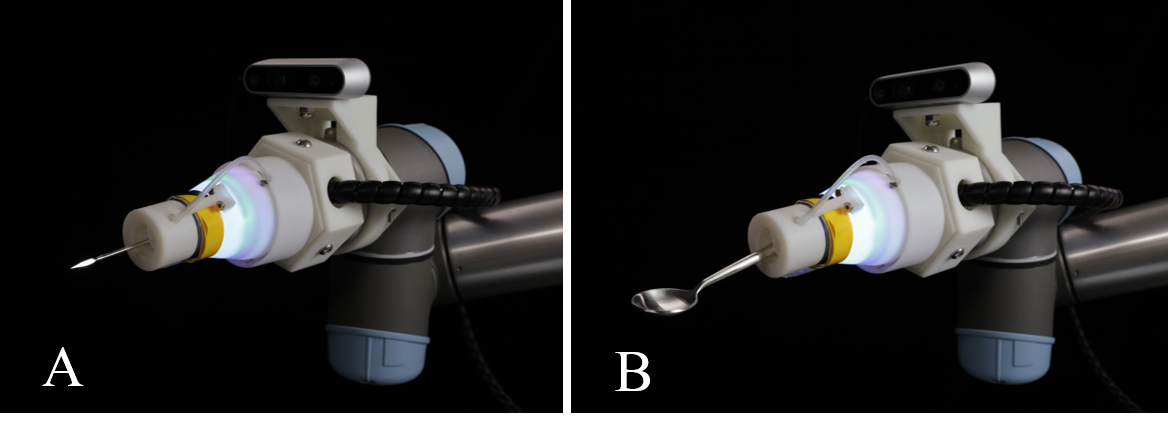}
	\caption{Other applications of the proposed device. (A) Scalpel gripping tool. (B) Feeding tools in senior care assistance. } \label{fig:Expandable}
\end{figure}

Considering the challenges and needs in the post-pandemic era, in this article, we design a novel OP swab sampling device with high compliance and low cost based on the visuo-tactile sensor and the soft pneumatic gripper, as shown in Fig.~\ref{fig:device}. The contributions of this article are trifold.
\begin{itemize}
\item The OP swab sampling device is designed by combining a concave structured visuo-tactile sensor named Cotac for high-resolution end force detection of the swabs and a rigidly wrapped pneumatic soft gripper of both strong gripping force and softness. The average force perception accuracy of the device can reach 0.052 N in X and Y directions.
\item Based on our sampling device, an adaptive virtual force-based (AVF) compliant control algorithm is proposed, which can accomplish the sampling task under the setting of task planning. In addition, an OP sampling position detection algorithm and a tactile information extraction algorithm are proposed.
\item Systematic and extensive experiments are designed and validate the compliance, safety, and comfort of the proposed OP swab sampling device and methods. The device can also be extended to other applications, including surgical robots, elderly assistance feeding robots, etc., as demonstrated in Fig.~\ref{fig:Expandable}.
\end{itemize}


\section{Structural Design}
\begin{figure*}
	\centering
	\includegraphics[width=1\textwidth]{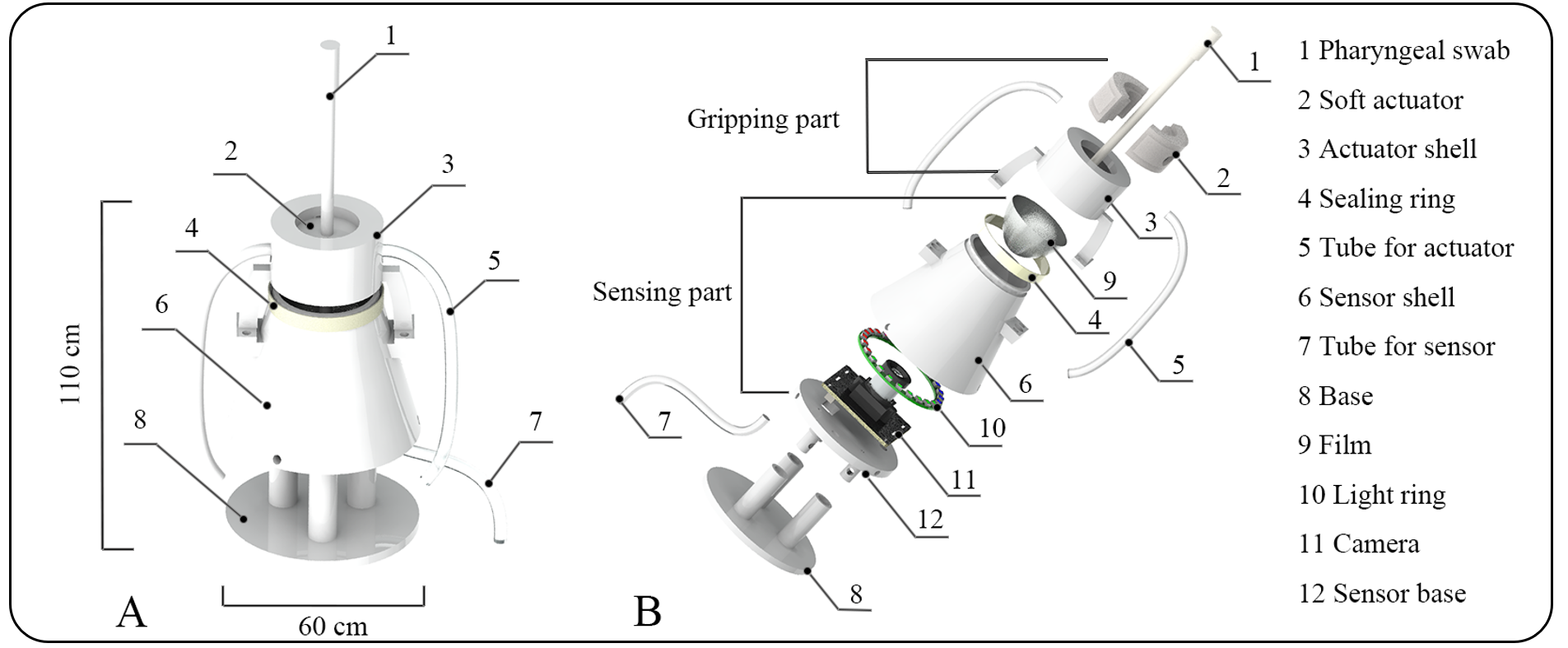}
	\caption{OP swab sampling device structure diagram. (A) Overall Structure. (B) Explosive View.} \label{fig:structure}
\end{figure*}

 According to the function, the device can be divided into two parts, i.e., the visuo-tactile sensing part and the pneumatic gripping part, which will be introduced in detail.


\subsection{Visuo-Tactile based Sensing Part}
To improve the tactile perception capability of the OP sampling devices, we present a concave structured tactile sensor named CoTac, which adopts a visuo-tactile sensing method with low cost and high resolution, and the structure is shown in Fig.~\ref{fig:structure}. Here, an ultra-thin elastic silicone film is utilized as the sensor surface, which could improve the sensitivity of the tactile sensor compared with the traditional acrylic and silicone solutions such as GelSight~\cite{yuan2017gelsight} and DIGIT~\cite{9811806}.

\begin{figure}
	\centering
	\includegraphics[width=0.48\textwidth]{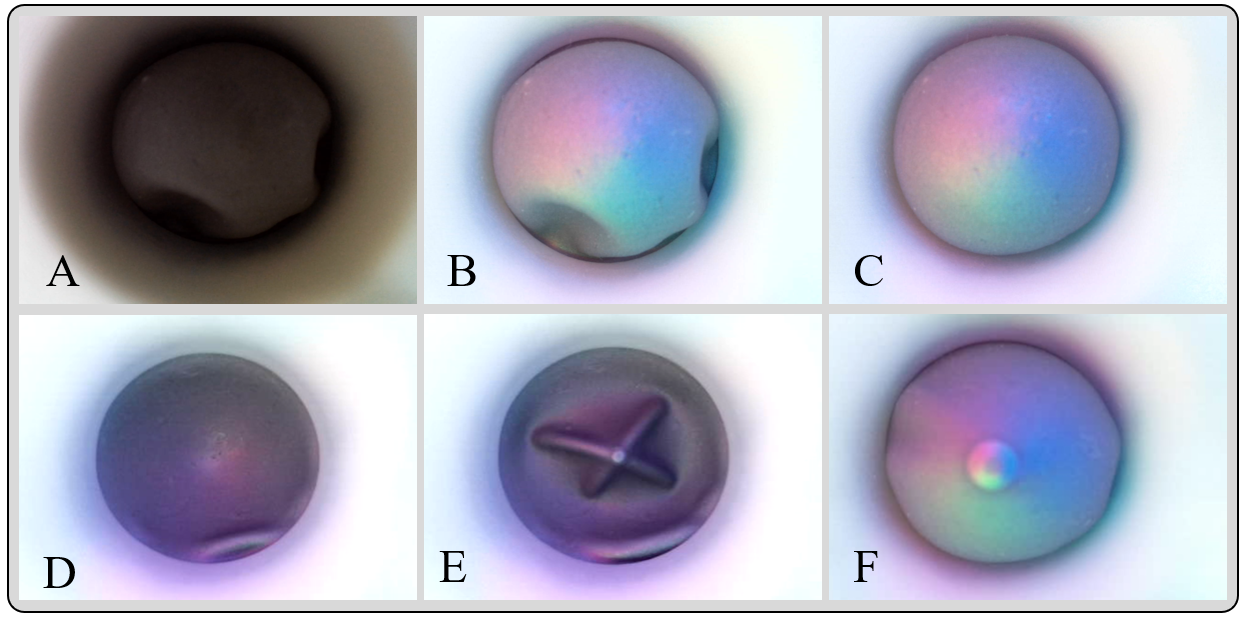}
	\caption{Tactile information in different conditions: (A) Under natural light conditions. (B) Under RGB light conditions. (C) Under negative pressure and RGB light conditions. (D) On convex surfaces under LED lighting. (E) When the swab contacts a convex surface. (F) When the swab contacts a concave surface. } \label{fig:Tactile}
\end{figure}

\begin{figure}
	\centering
	\includegraphics[width=0.48\textwidth]{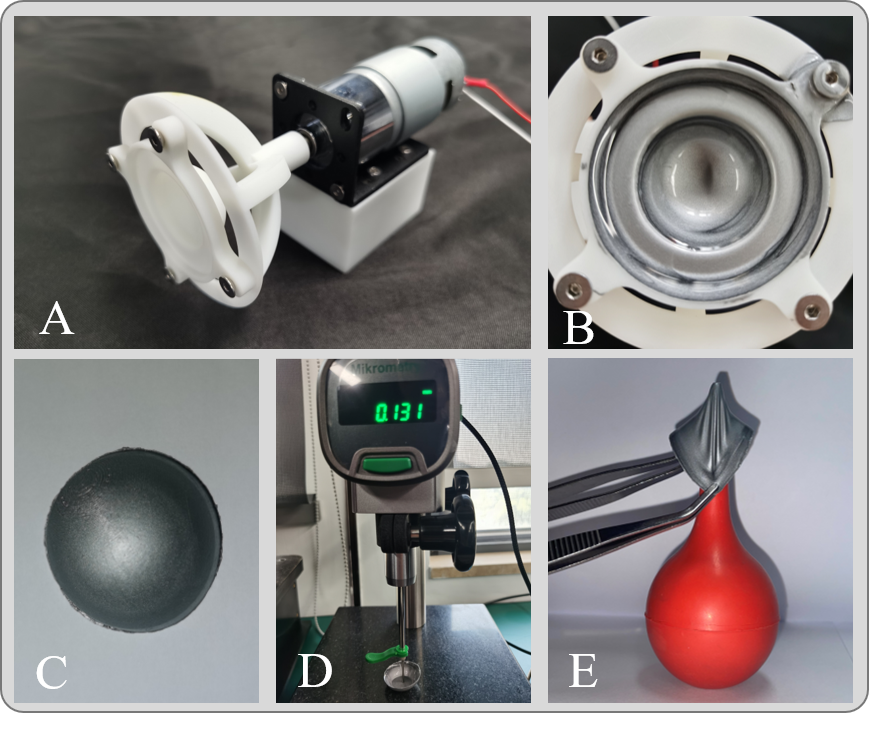}
	\caption{Sensor surface elastic film. (A) Production Tools. (B) Results after curing. (C) Film after cutting. (D)Thickness measurement. (E) Tensile performance test.} \label{fig:surface}
\end{figure}

Furthermore, to increase the perceptual effect, we use RGB light strips to illuminate the inspection surface and use an air pump to make the inside of it a negative pressure state so that the film is in a regular shape, the results are shown in Fig.~\ref{fig:Tactile}(A)(B)(C). Compared with the convex inflatable structure~\cite{kuppuswamy2020soft,alspach2019soft,li2021design}, the inner concave elastic film can have a better light effect and produce fewer folds when in contact with the OP swab, as depicted in Fig.~\ref{fig:Tactile}(D)(E)(F).
To fabricate the ultra-thin elastic film, we use Ecoflexx-50 as the production material, which has stronger toughness compared with Ecoflexx-10. To make the elastic film thickness uniform, a rotary silicone surface production process is designed, as illustrated in Fig.~\ref{fig:surface}, using centrifugal force to overcome gravity, by adjusting the speed to control the elastic film thickness. In this work, the fabricated elastic film has a thickness of 0.131 mm, and the tensile performance can reach 3 times its original length (300\% stretchability).

\subsection{Pneumatic Gripping Part}
The gripping part mimics the state of a doctor's fingers during sampling.  Based on our survey with the medical staff, we found that the human finger parallels the gripping structure and provides better stability and control accuracy. 
The reason why the human hand can grip the swab steadily is that the bones give muscle support, muscle provides power, skin provides a certain amount of friction, and the combination of the three can generate a strong grasping ability.
Inspired by this, a pneumatic soft swab gripping part with a rigid skeleton is proposed, as shown in Fig.~\ref{fig:structure}.

To analyze the effect of the number of actuators on the swab gripping, we performed finite element analysis on two actuators and three actuators. The analysis results proved that in a small space, three actuators of the structure before it easy to interfere with each other thus creating a gap in the middle, and the OP swab is thin, it is difficult to achieve effective gripping. The structure of the four actuators is more difficult to make in such a narrow space, and the stability of each actuator is not well guaranteed. Therefore, we finally chose the structure of two actuators.
\begin{figure*}
	\centering
	\includegraphics[width=0.96\textwidth]{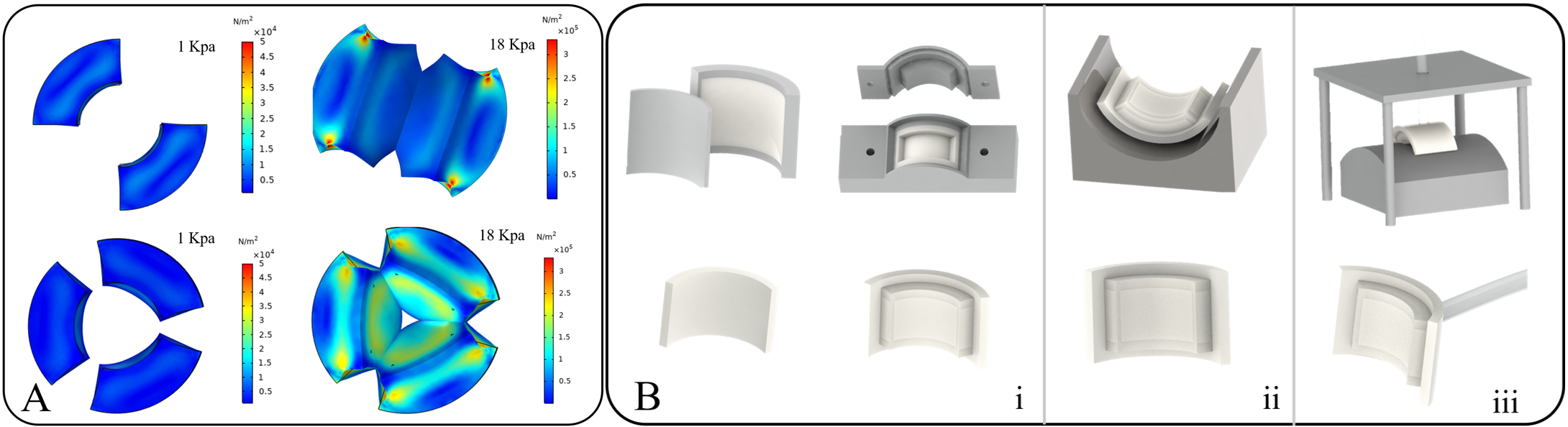}
	\caption{(A) Finite Element Analysis (FEA) using Comsol software. (B) Soft  actuator production process; (i) Make the top and bottom of the actuator; (ii) Stick the two parts together; (iii) Stick the air tube to the bottom of the actuator. } \label{fig:FEA}
\end{figure*}

For the fabrication process of the actuator, we adopted a step-by-step method, and the production process is shown in Fig.~\ref{fig:FEA}(B), the production material is Ecoflex-30, and the adhesive is Sil-Poxy.
Since the device does not contain any electronic components in addition to the built-in camera, it has good waterproof performance and can meet the needs of frequent spray disinfection during the sampling process.
\subsection{Device Cost}

To ensure the safety and reliability of the sampling process, swab sampling end-effectors often require precise feedback motors or multidimensional force sensors\cite{seo2020development,globaltimes}, which are expensive. Compared with the current pharyngeal swab sampling gripper, our proposed sampling device has a lower cost on the basis of ensuring safety and suppleness.
The device mainly includes one camera, two air pumps, one light strip, 3D printed parts, and silicone. In addition, we  make two small constant-pressure devices using plastic bottles at a cost of about 1 USD, which has a very good performance due to the low air pressure used in the device. As can be found in Table I, the total cost of the device is less than 30 USD and the weight is less than 0.1 kg.


\begin{table}[h]
 \centering
 \caption{Cost and Weight Details of the Proposed Device}
 \label{tab:pagenum}
 \begin{tabular}{l | c | c | c}
  \hline \hline
  & Number & Cost (USD)  & Weight (g) \\
  \hline
  Air pump (ZR520)  & 2   & 5  &    -   \\
  Camera (WSD-RY5M)  & 1 & 10   &  9   \\
  Light strip   & 1 & 3    &  1.5  \\
  3D printed parts   & - & 3    &    70   \\
  Silicone           & - & 1     &     3  \\
  Constant pressure device     & 2   & 0.5   &  -  \\
  \hline
  Total                    & -   & 28   &  83.5 \\
  \hline \hline
 \end{tabular}
\end{table}

\section{Algorithm Design}
\begin{figure}[ht]
	\centering
	\includegraphics[width=0.48\textwidth]{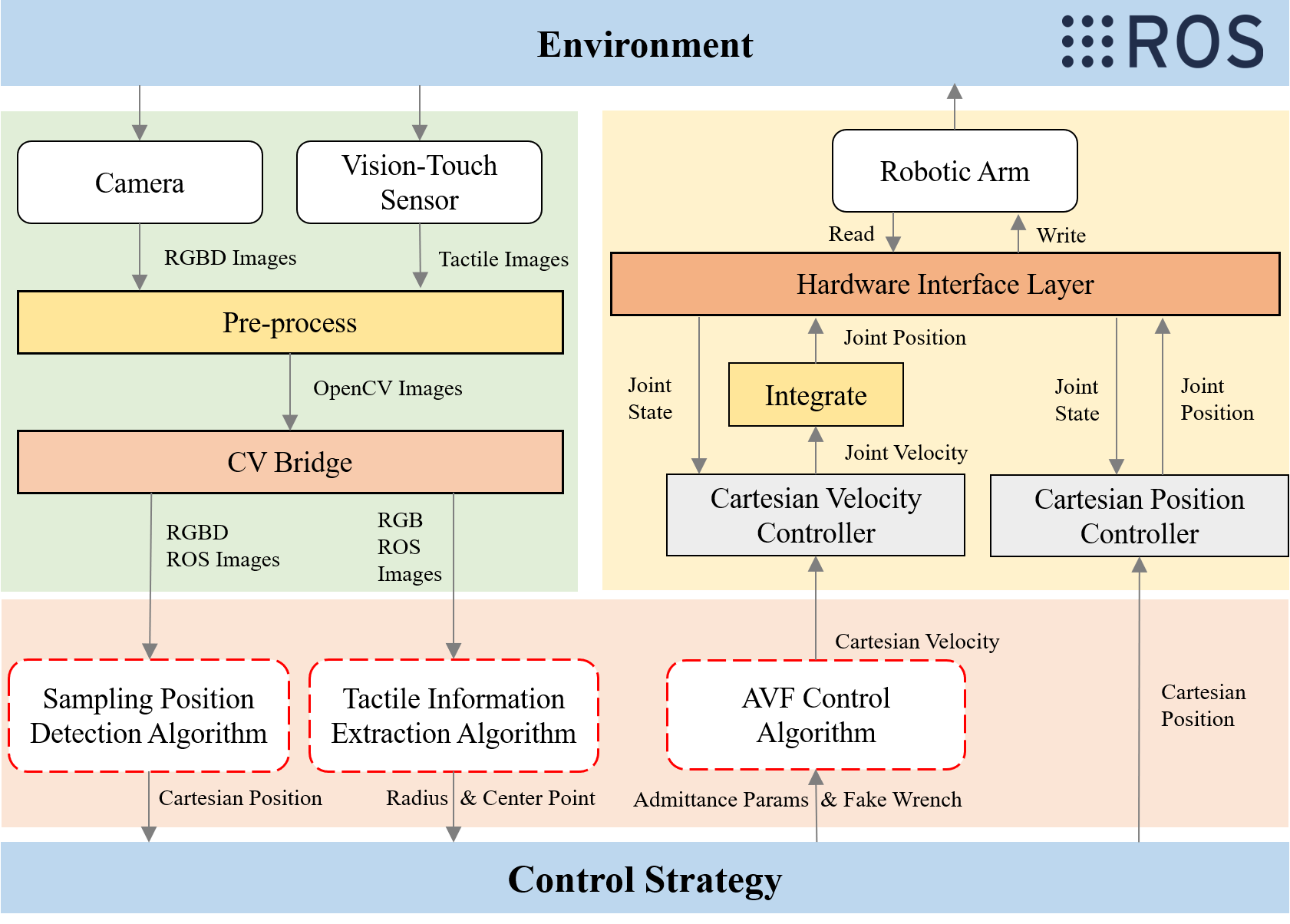}
	\caption{Framework of OP swab sampling algorithm based on Robot Operating System (ROS).} \label{fig:framework}
\end{figure}

How to balance comfort and effectiveness has been the focus of research on OP swab sampling robots. Due to the small space in the oral cavity and the obstruction of the hand, it is difficult to accurately determine the detection position based on vision. For the medical staff, a large part of the information is obtained through tactile sensation when performing the OP swab sampling. Therefore, we propose a robotic OP swab sampling method based on the CoTac sensor and admittance control to achieve safe and stable sampling. The method includes a sampling position detection algorithm, a tactile information extraction algorithm, and a compliant control algorithm, of which the flow chart is described in Fig.~\ref{fig:framework}. 


\subsection{Sampling Position Detection Algorithm}
\begin{figure}[htp]
	\centering
	\includegraphics[width=0.48\textwidth]{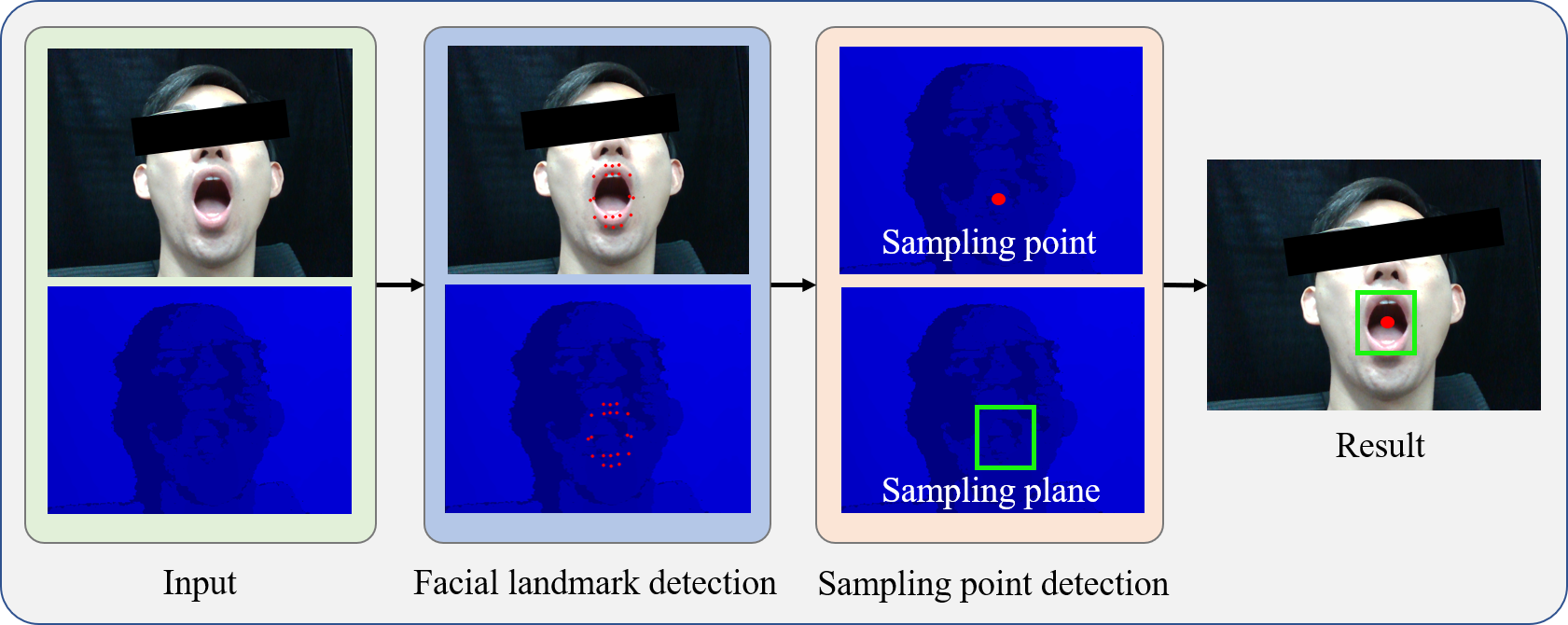}
	\caption{Schematic illustration of the sampling position detection algorithm.} \label{fig:position}
\end{figure}

The facial landmark detection algorithm~\cite{kazemi2014one} is a simple and efficient method to achieve accurate human feature point detection without the need for powerful computing power and a huge data set. Based on this algorithm, we propose a method for detecting oral sampling points in three dimensions, the algorithm process is shown in Fig.~\ref{fig:position}. First, the feature points of the lips are detected using the face alignment algorithm and the minimum outer rectangle of the region where the feature points are located is calculated, and then the feature points and the minimum outer rectangle are mapped to the depth image. Next, the deepest point in the rectangle is used as the sampling point. Finally, to determine the sampling direction, we fit a plane using the outermost feature points of the lips, and the normal vector of the plane is used as the direction for the OP swab when sampling. Once the sampling point and the sampling direction are obtained, the robot arm can be controlled to perform the sampling.

\begin{figure} [htp]
	\centering
	\includegraphics[width=0.48\textwidth]{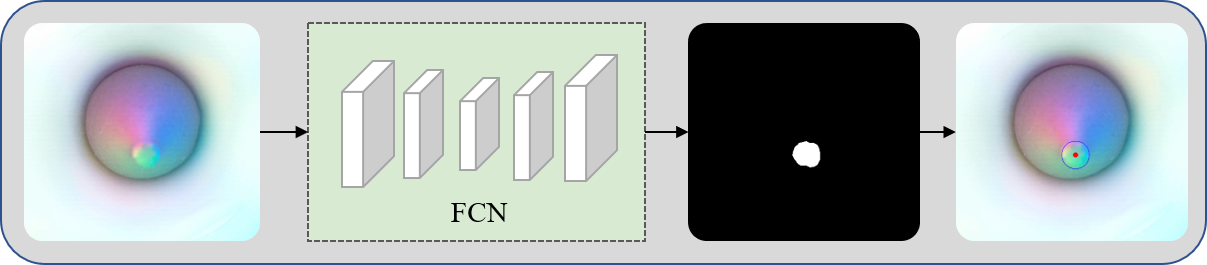}
	\caption{Tactile information extraction algorithm based on FCN.} \label{fig:Contact}
\end{figure}

\subsection{Tactile Information Extraction Algorithm}
Since the surface of the elastic film is curved, it is difficult to extract the contact information between the swab and the silicone surface effectively by traditional vision methods~\cite{long1990stationary}. To solve this problem, we use Fully Convolutional Networks (FCN)~\cite{long2015fully} for tactile information extraction and the process is illustrated in Fig.~\ref{fig:Contact}. We collected 200 images as the training data for network training, and after 60 rounds of iterations, the feature extraction accuracy reaches 99.2\%.

\subsection{AVF Control Algorithm}
In the process of sampling, we have used two controllers as shown in Fig.~\ref{fig:framework}. The pre-sampling process is completed in Cartesian position controller. The medical staff has high-precision perception and force control abilities, which can ensure sampling accuracy and well avoid the occurrence of sampling accidents. To avoid the discomfort or danger caused by unexpected operations of the robot during the sampling process, we propose an AVF control algorithm in Cartesian velocity controller. It can complete the sampling task safely with virtual force which is generated by the CoTac. This algorithm mainly consists of the admittance control algorithm \cite{7384497}, and adaptive proportional integral derivative (PID) control algorithm \cite{8287305},
\begin{equation}
    \mathbf{M}\Ddot{\mathbf{x}}(t)+\mathbf{D}\Dot{\mathbf{x}}(t)+\mathbf{S}(\mathbf{x}(t)-\mathbf{x}_d)=\mathbf{f}(t),
\end{equation}
where $\mathbf{M}$, $\mathbf{D}$ and $\mathbf{S} \in \mathbb{R}^{3\times3}$ denote the mass, damping and stiffness matrices, respectively. $\Ddot{\mathbf{x}}(t)$, $\Dot{\mathbf{x}}(t)$, $\mathbf{x}(t)$, $\mathbf{x}_d$, and $\mathbf{f}(t) \in \mathbb{R} ^3 $ are the end states of the robotic arm which are comprised of the acceleration, velocity, position, desired position and virtual external force.
\begin{equation}
    \mathbf{f}(t) = \mathbf{f}_{o} + \mathbf{K}_P\mathbf{e}(t) + \mathbf{K}_I\int_0^T \mathbf{e}(t)dt +\mathbf{K}_D\frac{d(\mathbf{e}(t))}{dt},
\end{equation}
where $\mathbf{e}(t) = \mathbf{c}_c(t) - \mathbf{c}_d(t)$ and $\mathbf{c}_c(t)$, $ \mathbf{c}_d(t) \in \mathbb{R}^3$ represent the position and radius of the center point and desired point in the sensor image. $\mathbf{K}_p$, $\mathbf{K}_i$ and $\mathbf{K}_d \in \mathbb{R}^{3\times3}$ are the adaptive PID parameters. $\mathbf{f}_{o} \in \mathbb{R}^3$ is the virtual force offset which can be set to change the sampling directions. In the real system deployment, we adopt the incremental PID control algorithm and adaptive function for discrete state control, which are given by:
\begin{equation}
    {\mathbf{f}(n)} = {\mathbf{f}(n-1)} + \mathbf{K}\left[\Sigma_{i=1}^3 \mathbf{w}_i(n) \odot  \mathbf{z}_i(n)\right],
\end{equation}
\begin{equation}
    \mathbf{w}_i(n) = \mathbf{w}_i(n-1)+\mathbf{v}_i(n)\odot\left[\mathbf{u}(n)-\mathbf{f}(n)\right]\odot\mathbf{f}(n)\odot \mathbf{z}_i(n),
\end{equation}
where $\odot$ denotes the element-wise product. Adaptive neural network  has been used to adjust the PID parameters online \cite{8287305}. $\mathbf{w}_i$ is the weight matrix and $\mathbf{z}_1(n) = \mathbf{e}(n) - \mathbf{e}(n-1)$, $\mathbf{z}_2(n) = \mathbf{e}(n)$, $\mathbf{z}_3(n) = \mathbf{e}(n)-2\mathbf{e}(n-1)+\mathbf{e}(n-2)$. $\mathbf{v}_i(n) \in \mathbb{R}^{3}$ is the learning rate of the adaptive function and $\mathbf{u}(n) \in \mathbb{R}^{3}$ is the desired output. Compared with a fixed virtual external force set, the adaptive function has been used to improve the system robustness. The output end velocity of the robotic arm can be solved on the basis of the above equations in one control cycle:
\begin{equation}
    \Ddot{\mathbf{x}}(n) = \mathbf{M}^{-1}\left\{\mathbf{f}(n)-\left[\mathbf{D} \dot{\mathbf{x}}(n) + \mathbf{S}(\mathbf{x}(n)-\mathbf{x}_d)\right] \right\},
\end{equation}    
\begin{equation}
    \dot{\mathbf{x}}(n+1) = \dot{\mathbf{x}}(n) + \Ddot{\mathbf{x}}(n)\Delta{t},
\end{equation}
where $\Delta{t} = 8$~ms, because of the hardware control cycle.

\begin{figure}[htp]
	\centering
	\includegraphics[width=0.48\textwidth]{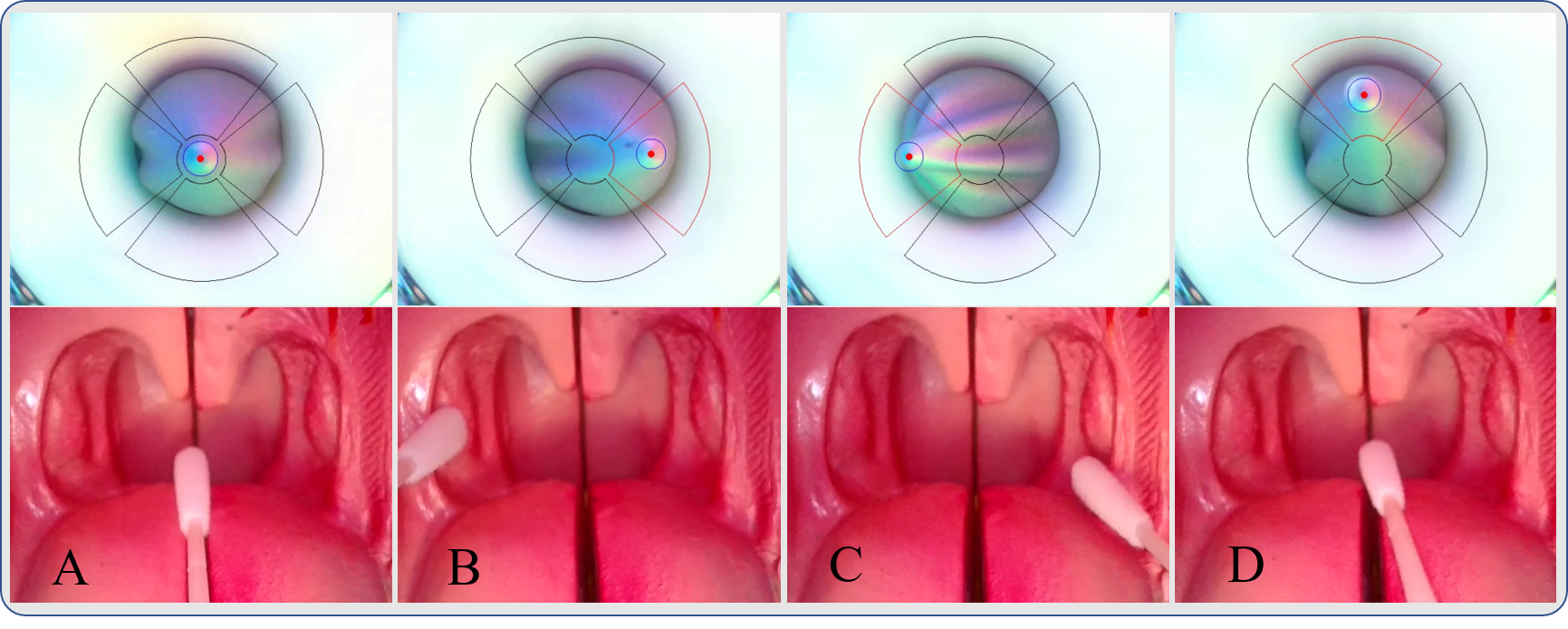}
	\caption{Sampling process with tactile feedback. (A) Initial position. (B) Pharyngeal left sampling. (C) Pharyngeal right sampling. (D) Pharyngeal middle sampling.} \label{fig:feedback}
\end{figure}

During OP swab sampling, the swab will move from the desired position to three sampling areas, including the left, right, and middle, as depicted in Fig.~\ref{fig:feedback}. When it touches the inner wall of the oral cavity, the end of the sampling swab will produce elastic deformation in the sensor. Then the offset will be recognized by the tactile information extraction algorithm and used as the virtual external force to complete the sampling task.

\IncMargin{1em}
\begin{algorithm} \SetKwData{Left}{left}\SetKwData{This}{this}\SetKwData{Up}{up}  \SetKwInOut{Input}{input}\SetKwInOut{Output}{output}
	
\Input{Sampling times $L$, current end states of the robotic arm $\mathbf{x},\dot{\mathbf{x}}$, desired end state of the robotic arm $\mathbf{x}_d$, virtual force set $\mathbf{f}_0$, virtual external force $\mathbf{f}$, PID parameters, learning rates $\mathbf{v}_i$ and error threshold $\epsilon$} 
\Output{Next end state of the robotic arm $\dot{\mathbf{x}}$}
\BlankLine 

\For{$l\leftarrow 1$ \KwTo $L$}
{
$j\leftarrow 1$\\
\While{$\Vert\mathbf{M}\ddot{\mathbf{x}}(j)+\mathbf{D} \dot{\mathbf{x}}(j)+\mathbf{S}(\mathbf{x}(j)-\mathbf{x}_d) - \mathbf{f}(j)\Vert$ $\geq \epsilon$}{
\Repeat{this end condition}{
Calculate and output the next moment velocity $\dot{\mathbf{x}}(j+1)=\dot{\mathbf{x}}(j)+\ddot{\mathbf{x}}(j)\Delta t$\;
Update the robot states $\mathbf{x},\dot{\mathbf{x}}$ \;
Update adaptive PID parameters $\mathbf{w}_i(j+1) = \mathbf{w}_i(j)+\mathbf{v}_i(j)\odot\left[\mathbf{u}(j)-\mathbf{f}(j)\right]\odot\mathbf{f}(j)\odot \mathbf{z}_i(j), i=\{1,2,3\}$\;
Update the virtual external force 
$\mathbf{f}(j+1) = \mathbf{f}(j) + \mathbf{K} \left[\Sigma_{i=1}^3 \mathbf{w}_i(j) \odot  \mathbf{z}_i(j)\right]$\;
}
$j = j+1$\;}
$m=m+1$\;
\emph{Update virtual force $\mathbf{f}$}\;}
\caption{Sampling System with Adaptive Compliant Control Algorithm}
\end{algorithm}
\DecMargin{1em}

All the above functions are implemented under the ROS platform which makes the deployment to be completed conveniently. The servo control cycle of the robotic arm matches the control decision cycle, and two kinds of controllers (velocity and position) can be easily switched in this project.


\section{Experiment Evaluation}
To test the effectiveness of the proposed OP swab sampling device, we design a series of experiments in a systematic manner, including the gripping force experiment, swab gripping experiment, force compliance experiment, sampling effectiveness experiment, and in vivo experiment.

\subsection{Gripping Force Experiment}
\begin{figure}
	\centering
	\includegraphics[width=0.48\textwidth]{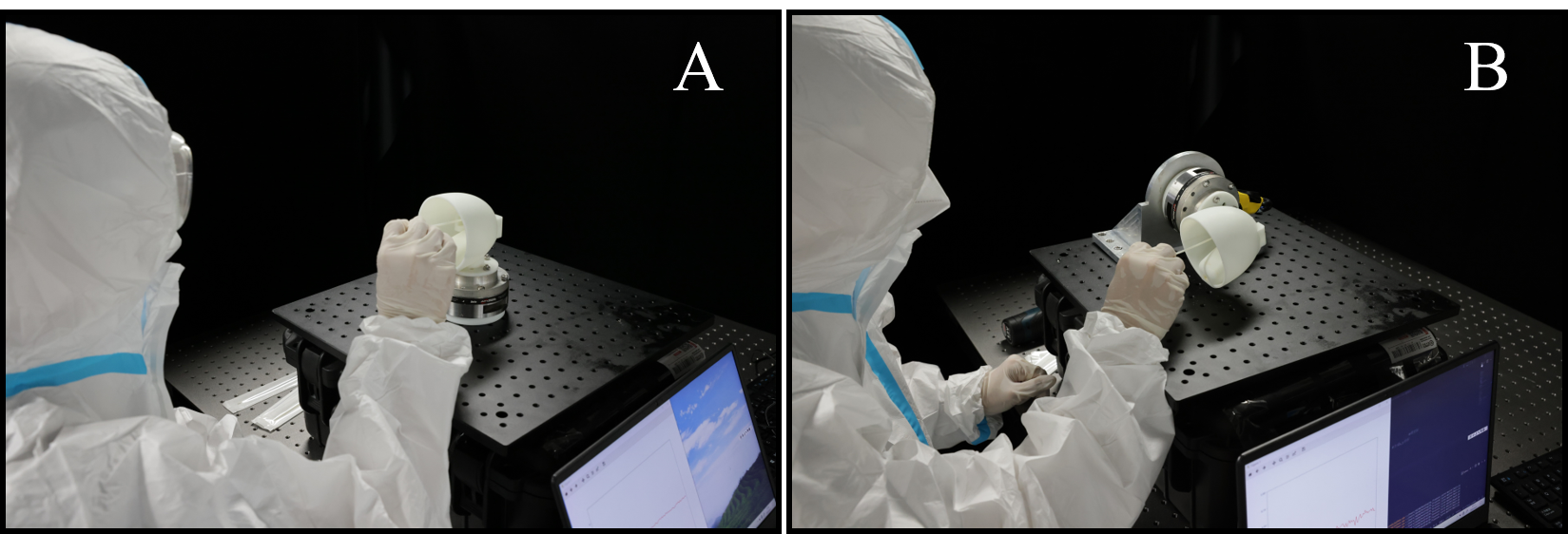}
	\caption{Real sampling data collection. (A) X-axis force test. (B) Y-axis force test. } \label{fig:real_test}
\end{figure}

\begin{figure}
	\centering
	\includegraphics[width=0.48\textwidth]{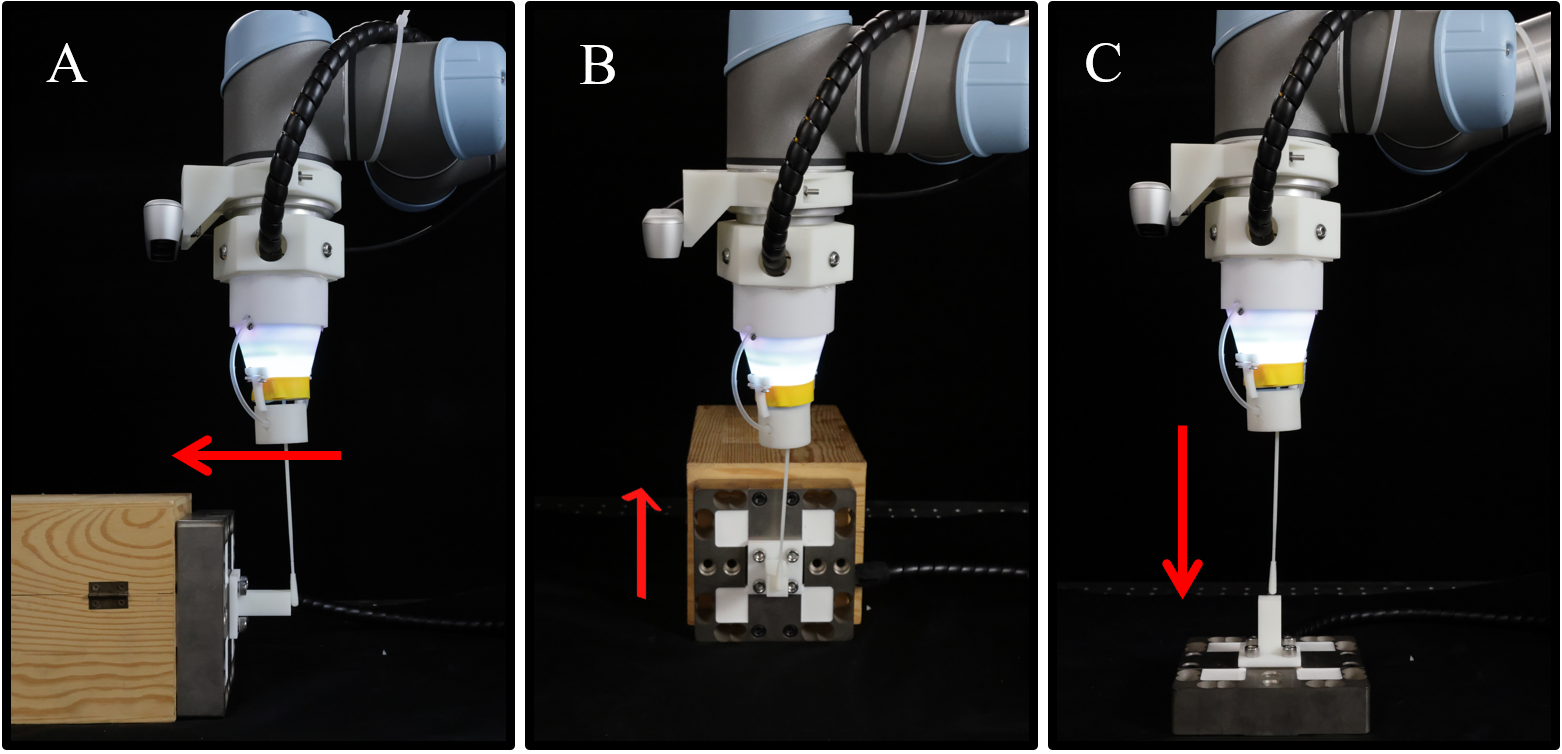}
	\caption{Force test experiment. (A) X-axis force test; (B) Y-axis force test; (C) Z-axis force test. } \label{fig:Force}
\end{figure}

A swab sampling force experiment is performed to test if the gripping structure can provide sufficient force and is safe enough for OP swab sampling. As a preliminary, we measured the sampling force provided by the medical staff as shown in Fig.~\ref{fig:real_test}. It shows that the force exerted by the medical staff at the end of the OP swabs in the X and Y axes during OP swabs sampling is about 0.15~0.3 N.
This number will be used as a reference for our robotic sampling. For our sampling device, the contact force is mainly determined by the grasping pressure and the position offset. To figure out the quantitative relationship, we use the DYDW-005 force sensor to measure the contact force of the sampling device in three axes, as demonstrated in Fig.~\ref{fig:Force}. 

\begin{figure}[ht]
	\centering
	\includegraphics[width=0.48\textwidth]{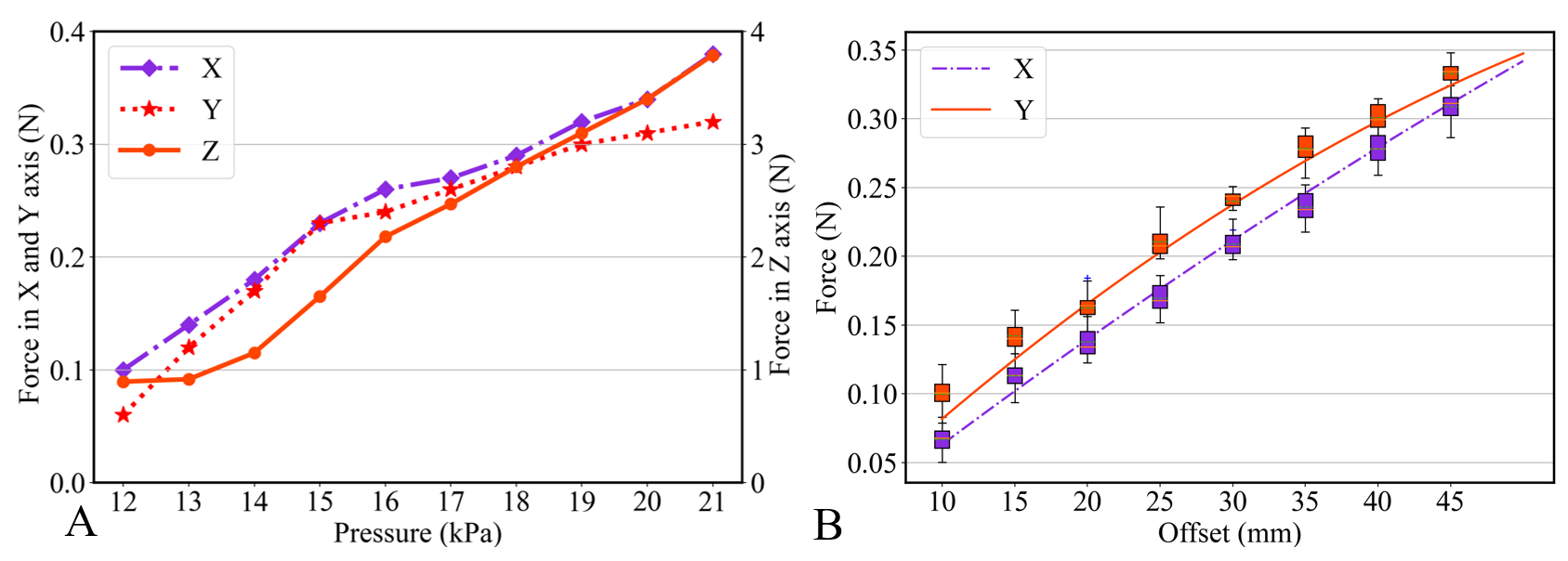}
	\caption{Experimental results. (A) Relationship between air pressure and force. (B) Relationship between offset and force.} \label{fig:relationship}
\end{figure}

First, we fix the offset and change the gripping pressure. For the Z-axis contact force test, we control the device to drop 1.5 cm. While for X-axis and Y-axis force tests, the end of the swab is offset by 4 cm. For each test, we record the corresponding gripping force under different air pressures. The resulting contact force is shown in Fig.~\ref{fig:relationship}(A). It can be observed that the contact force varies approximately linearly with the air pressure, therefore, by adjusting the air pressure, we can get the appropriate gripping force.


\begin{figure}
	\centering
	\includegraphics[width=0.48\textwidth]{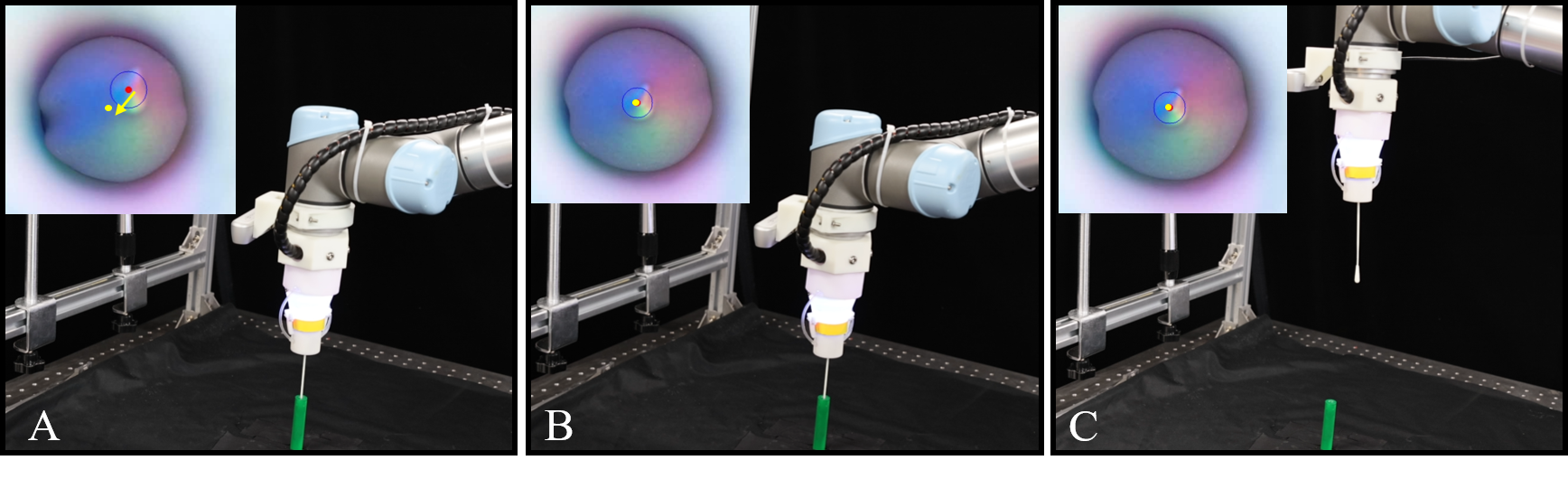}
	\caption{OP swab gripping process. (A) Deviation in the initial gripping position. (B) Adjusting the gripping position. (B) Finishing the gripping. } \label{fig:gripping}
\end{figure}

The significance of the Z-axis gripping force is to ensure that the  swab will not slide during the sampling process, when the Z-axis offset is too large, the swab will appear passive bending, so the test offset on the Z-axis gripping force is not significant. When the air pressure is greater than 16 kPa, it can withstand more than 2 N force in the Z axis to meet the requirement of the swab gripping force. X-axis and Y-axis forces are the keys to achieving compliance. Since the sampling force of the medical staff is about 0.15-0.3 N, the gripping force will be in a suitable range when the air pressure is 19 kPa.
 
Next, to get the most suitable offset, we tested the contact force at the end of the OP swab for different offsets at a fixed pressure of 19 kPa.
To obtain the relationship between force and offset of the sampling device, we collected 160 sets of data with an offset between 10-45mm and used the quadratic equation to establish the relationship between offset and force, the obtained results are shown in Fig.~\ref{fig:relationship}(B). Through the test, the average force perception accuracy of the device in 0.05N~0.35N intervals can reach 0.052 N in X and Y directions.

\subsection{Swab Gripping Experiment}
The Cotac sensor not only has higher stability and lower cost compared to traditional tactile sensors but also has a larger detection area and higher resolution. The concave detection surface plays an important role in swab gripping. In each gripping, although the swab is placed in a fixed initial position, there will inevitably be deviations in the position, which will have an impact on the gripping and sampling results. Therefore, we propose a tactile calibration-based OP swab gripping method to improve the success rate and consistency of detection by adjusting the gripping position according to tactile information when gripping the swab, so that the swab is always kept in the center of the device when gripping finishes. The process is shown in Fig.~\ref{fig:gripping}.

\subsection{Force Compliance Experiment}
\begin{figure*}
	\centering
	\includegraphics[width=0.96\textwidth]{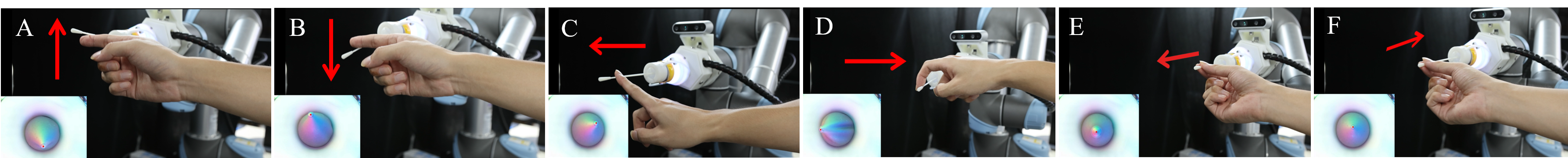}
	\caption{Force compliance experiment. (A) The upward force on the swab. (B) The downward force on the swab. (C) The leftward force on the swab. (D) The rightward force on the swab. (E) The backward force on the swab. (F) The forward force on the swab.} \label{fig:compliant}
\end{figure*}

To test the performance of the sensor and the force compliance algorithm, we designed a hand-touching experiment as shown in Fig.~\ref{fig:compliant}. In this experiment, a person touches the end of the swab from different directions and the device can feel and follow the motion of the person's hand. The touching force in the X and Y axes is measured by the deviation of the contact position from the initial contact position, and the force in the Z axis is obtained from the diameter of the contact point. The same force compliance algorithm adopted in the OP swab sampling is applied here based on the tactile information.

The results show that the system has a high sensitivity to the end contact force, even a slight force on the end of the OP swab sampling device can achieve a good force compliance effect, and a single finger can easily push the robot arm. Specifically, the device can react to a touching force as small as 0.05 N, which verifies its excellent force compliance capability.

\begin{figure*}
	\centering
	\includegraphics[width=0.96\textwidth]{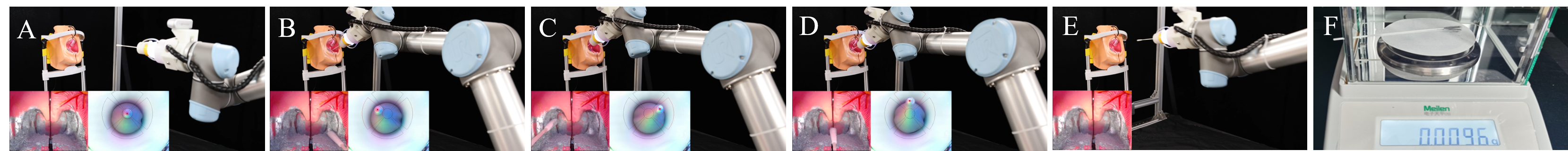}
	\caption{{Sampling effectiveness experiment. (A) Initial position. (B) Pharyngeal left sampling. (C) Pharyngeal right sampling. (D) Pharyngeal down sampling. (E) Complete sampling. (F) Measuring swab mass change.}}\label{fig:Sampling}
\end{figure*}

\subsection{Sampling Effectiveness Experiment}
To test the effectiveness of the sampling device, we take sampling experiments on an oral cavity model, and the sampling process is shown in Fig.~\ref{fig:Sampling}. The throat of the model is covered with silver powder instead of oral saliva. The swab will be stained with silver powder during sampling. Therefore we can tell if the sampling is successful by weighing the mass of the swab before and after the test. In each test, we slightly adjusted the position of the model to simulate the deviation of visual detection. Since the density of the silver powder is very low, we used an electronic scale with an accuracy of 0.0001 g for measuring, and the sampling is considered valid when the mass increases by more than 0.006 g (the average sampling quality of medical staff under the same conditions). We conducted a total of 40 sampling experiments, and the efficiency rate was 100\%. The sampling duration can be controlled within 25 s, and the sampling speed can be further improved by increasing the movement speed of the robot arm and simplifying the sampling process.

\subsection{Waterproof Performance Experiment}

\begin{figure}[ht]
	\centering
	\includegraphics[width=0.48\textwidth]{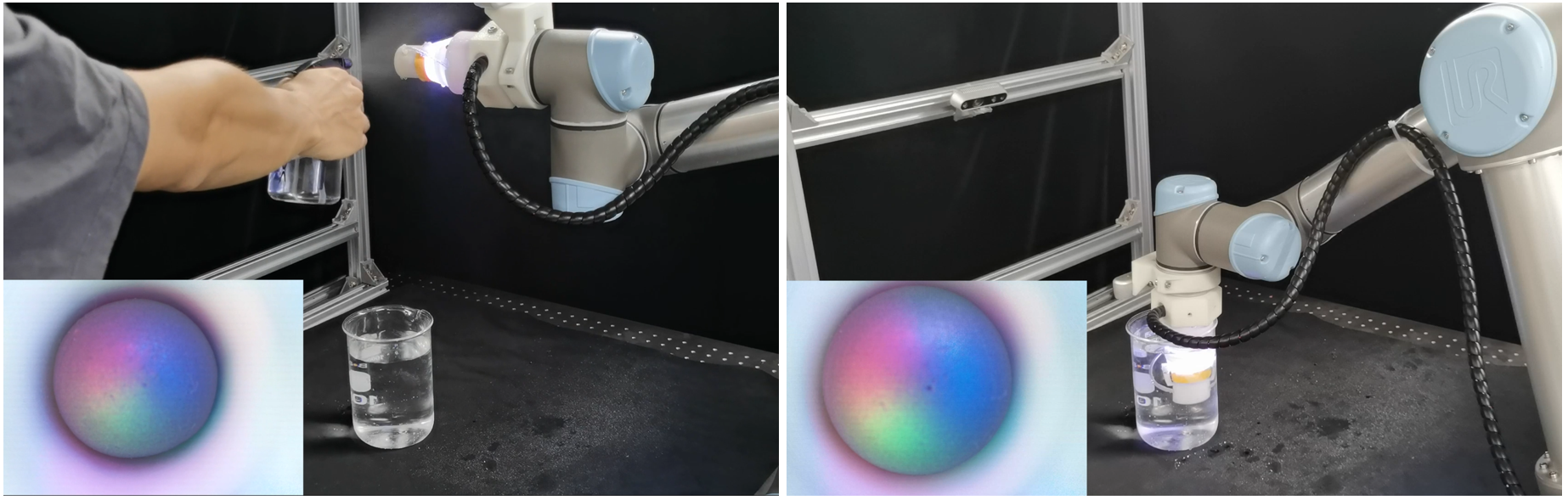}
	\caption{Waterproof Performance Experiment. (a) Alcohol spraying test.  (b) Alcohol immersion test. } \label{fig:Waterproof}
\end{figure}

To prevent cross-contamination, the equipment needs to be disinfected after each sampling, and the standard disinfection method is alcohol spraying, as shown in Fig.~\ref{fig:Waterproof}. To test the waterproof performance of the device, we conducted 1000 alcohol immersion experiments, each time immersed in alcohol for 10 s. After completing the tests, the gripping and sensing parts of the device still worked normally, mainly due to our pneumatic actuators and visual-tactile sensing technology.

\subsection{In Vivo Experiment}
\begin{figure}[ht]
	\centering
	\includegraphics[width=0.48\textwidth]{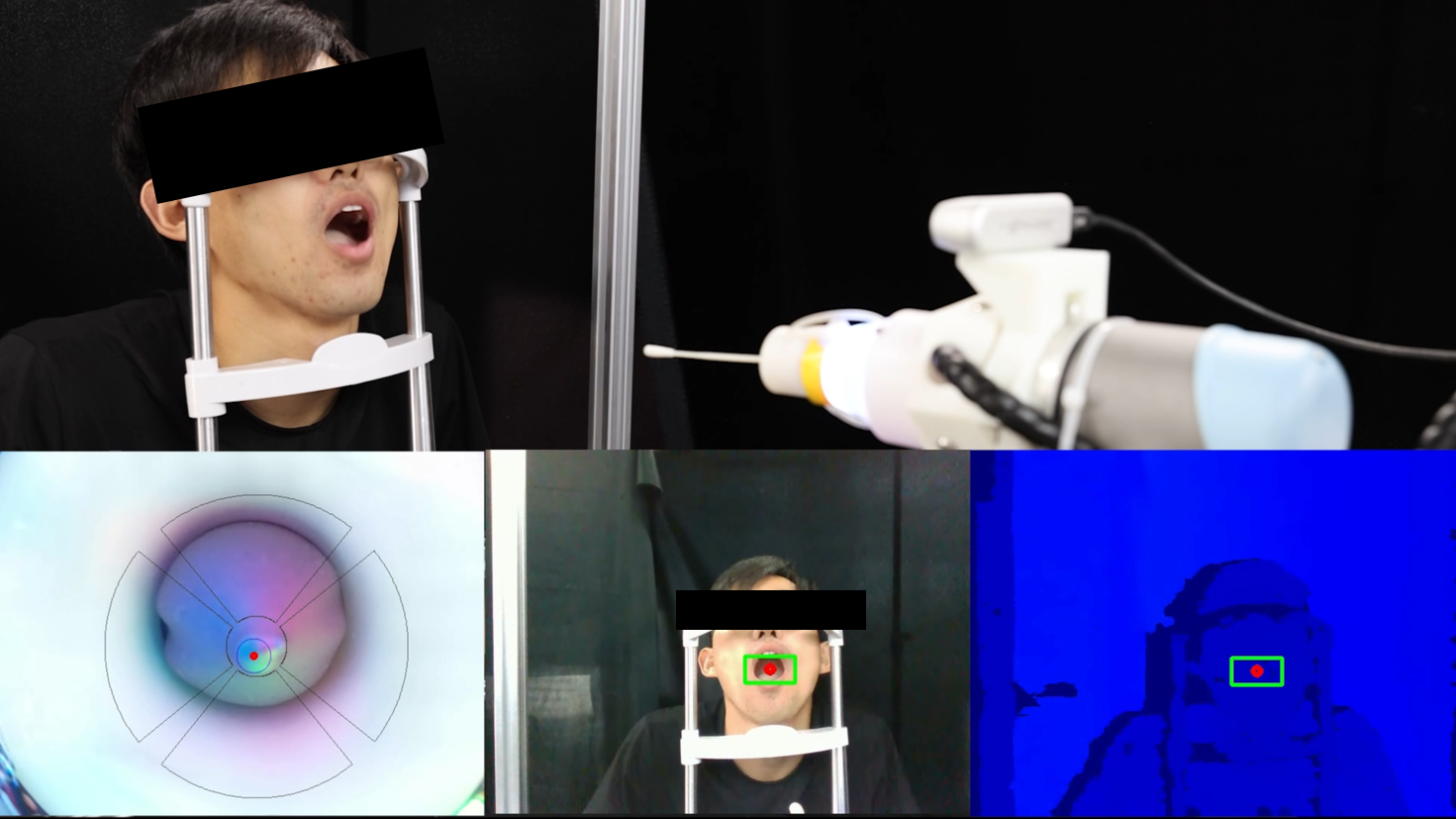}
	\caption{In vivo test experiment} \label{fig:vivo}
\end{figure}

To test the safety and comfort of the OP swab device in a real situation, we conducted 10 in vivo sampling tests, as shown in Fig.~\ref{fig:vivo}. To guarantee safety, the experiments were conducted under the supervision of professionals, and have asked for volunteers' permission. After the experiments, We conduct research on the comfort, safety, and effectiveness of each volunteer who takes part in the tests, the final evaluation scores (the total score for each item is 10)  are 9.1 for safety, 9.2 for comfort, 8.7 for sampling speed, and 9.5 for appearance. The experiment is only to compare the difference between human sampling and robot sampling and to provide a reference for the next improvement, so the experimental setup is relatively simple.

\section{Conclusion}
In this article, we propose a low-cost OP swab sampling device. In the sensing part, we propose a novel concave tactile sensor named CoTac, which has a large detection area as well as high sensitivity compared to the traditional force sensors. The device has a rigidly wrapped pneumatic soft structure in the gripping part, which not only generates an effective gripping force but also has high compliance, eliminating the potential threat to human safety. In addition, since the structure does not have an electric actuator, the whole device can be directly sprayed for disinfection to prevent cross-infection. For the sampling method, an OP sampling position detection algorithm, a tactile information extraction algorithm, and an AVF compliant control algorithm are proposed. Overall, we use both hardware compliance and algorithmic compliance to ensure the safety of the sampling process. The effectiveness of the device has been verified through sampling experiments as well as in vivo tests, indicating great application potential. The cost of the device is around 30 USD and the total weight of the functional part is 0.1 kg, allowing the device to be rapidly deployed on various robotic arms.

Currently, we mainly explore the performance of our proposed pharyngeal swab sampling device in terms of grasping and perception, and later we will further optimize the pharyngeal swab sampling system and add more research on sampling stability, sampling speed, and success rate.
\bibliographystyle{ieeetr}
\bibliography{refe}
\end{document}